\documentclass[11pt]{article}
\usepackage[final]{acl}
\usepackage{times}
\usepackage{latexsym}
\usepackage{graphicx}
\usepackage{booktabs}
\usepackage{multirow}
\usepackage{adjustbox}
\usepackage{algorithm}
\usepackage[noend]{algpseudocode}
\usepackage{fancyvrb}
\usepackage{framed}  
\usepackage{amsmath}
\usepackage{setspace} 
\usepackage{float}
\usepackage{amssymb}
\usepackage{enumitem}
\usepackage[utf8]{inputenc}
\usepackage{framed}
\usepackage{array}

\title{\textsc{CRAFT}: Training-Free Cascaded Retrieval for Tabular QA }

\author{$^\textbf{*}$Adarsh Singh$^{1}$ \quad $^\textbf{*}$Kushal Raj Bhandari$^{2}$ \quad  Jianxi Gao$^{2}$ \\ \quad \textsuperscript{\textdagger}\textbf{Soham Dan}$^{3}$ \quad \textsuperscript{\textdagger}\textbf{Vivek Gupta}$^{1}$ \\ 
        $^{1}$Arizona State University $^{2}$Rensselaer Polytechnic Institute  $^{3}$Microsoft \\
        \texttt{asing725@asu.edu}\quad \texttt{bhandk@rpi.edu}
        }
\begin{document}
    \maketitle
    \begingroup\def\thefootnote{}\footnotetext{$\textbf{*}$These authors contributed equally to this work. \quad \quad \textdagger These authors jointly supervised this work.}\endgroup
    \begin{abstract}
Open-Domain Table Question Answering (TQA) involves retrieving relevant tables from a large corpus to answer natural language queries. Traditional dense retrieval models, such as DTR and DPR, not only incur high computational costs for large-scale retrieval tasks but also require retraining or fine-tuning on new datasets, limiting their adaptability to evolving domains and knowledge. In this work, we propose $\textbf{CRAFT}$, a zero-shot, cascaded retrieval approach that first uses a sparse retrieval model to filter a subset of candidate tables before applying more computationally expensive dense models as re-rankers. 
To improve retrieval quality, we enrich table representations with descriptive titles and summaries generated by \textit{Gemini Flash 1.5}, enabling richer semantic matching between queries and tabular structures. Our method outperforms state-of-the-art (SOTA) sparse, dense, and hybrid retrievers on the NQ-Tables dataset. It also demonstrates strong zero-shot performance on the more challenging OTT-QA benchmark, achieving competitive results at higher recall thresholds, where the task requires multi-hop reasoning across both textual passages and relational tables. 
This work establishes a scalable and adaptable paradigm for table retrieval, bridging the gap between fine-tuned architectures and lightweight, plug-and-play retrieval systems. Code and data are available at:  \href{https://coral-lab-asu.github.io/CRAFT/}{https://coral-lab-asu.github.io/CRAFT/}


\end{abstract}
    \section{Introduction}

Tables serve as a vital format for representing structured knowledge, yet enabling open-domain question answering (QA) systems to retrieve and interpret tabular information effectively remains a non-trivial challenge. Unlike traditional Tabular QA (TQA) settings that assume access to gold tables, open-domain QA must first identify relevant tables from a large corpus before reasoning over their structure to derive accurate answers \cite{liangOpenDomainQuestionAnswering2024}. Simple approaches that linearize tables into text and apply standard text-retrieval methods can lose the syntactic and structural information encoded in rows and columns \cite{badaroTransformersTabularData2023, minExploringImpactTabletoText2024}. As a result, effective table retrieval has been identified as a key bottleneck for scalable QA over tabular data \cite{liTailoringTableRetrieval2025}. 
\begin{figure}[t]
    \centering
              \includegraphics[width=0.95\columnwidth]{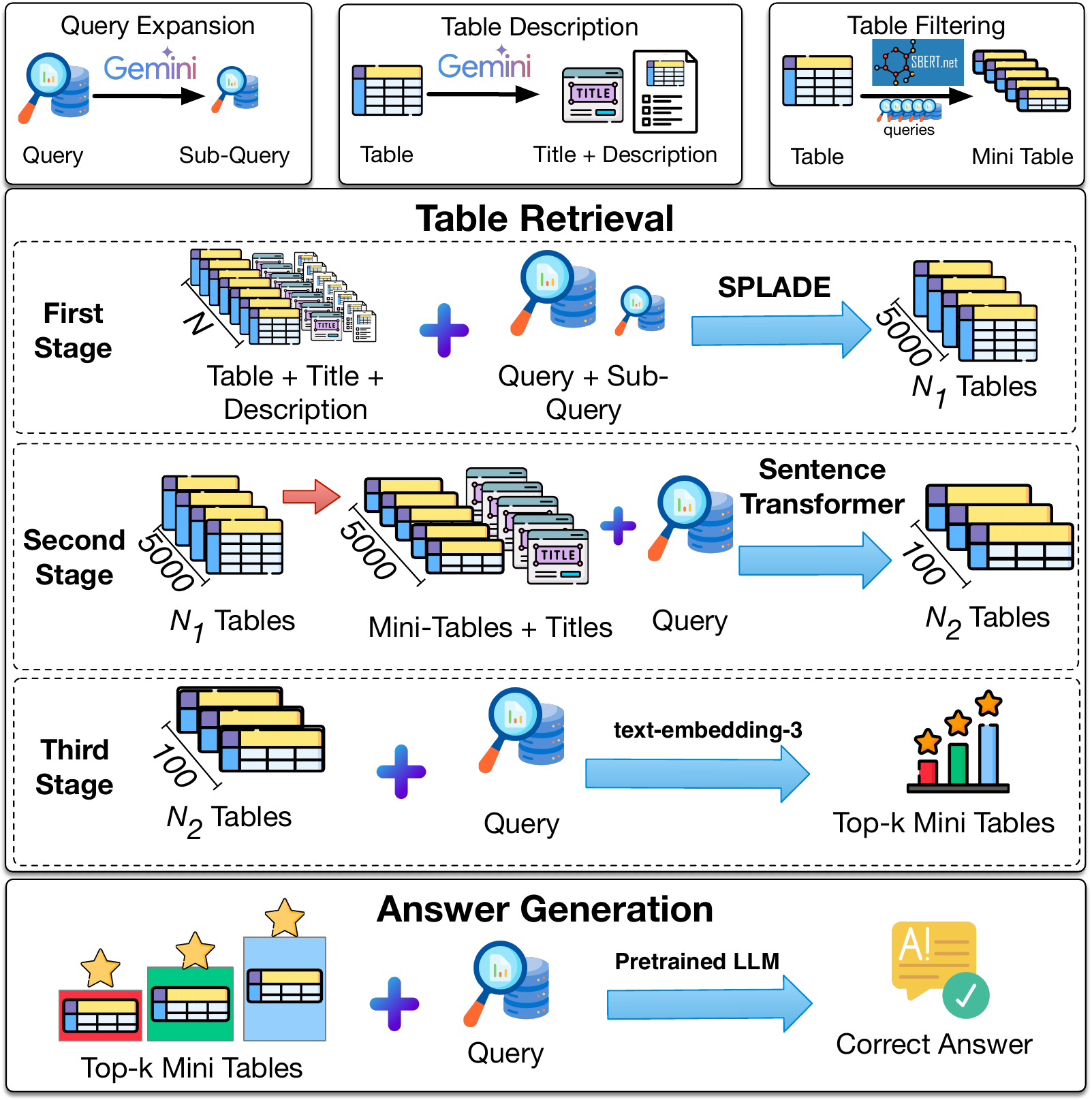}
              \caption{\small Overview of the \textbf{CRAFT} Framework.\vspace{-\baselineskip}}
              \label{fig:tablerag}
    \end{figure} 
    
Recent work has started to tackle open-domain TQA by developing specialized retrieval models. Early approaches adopted sparse or dense text retrievers on flattened tables (e.g., BM25 or bi-encoder models) \cite{fanSurveyRAGMeeting2024,robertsonProbabilisticRelevanceFramework2009}. The Dense Table Retriever (DTR) introduced by \citet{herzigOpenDomainQuestion2021} augments a bi-encoder with row and column features to better capture table structure. Several methods employ multi-stage cascades to prune candidates before applying costly scoring, with each stage acting as a separate ranking component. For example, ~\cite{wang2011cascade} introduced a two-stage document ranking cascade composed of a candidate generation step and a learned re-ranker, which are trained jointly as a single model. In a similar vein, the "Retrieve, Read, Rerank" framework from ~\cite{hu2019retrieve} embeds retrieval, answer extraction, and re-ranking within a single shared Transformer network that is fine-tuned end-to-end. The Re²G system ~\cite{glass-etal-2022-re2g} implements a retrieve-rerank-generate pipeline where the re-ranking and generation stages are merged into one BART-based sequence-to-sequence model fine-tuned with knowledge distillation.

More recent methods explicitly incorporate fine-grained structure. \citet{jinEnhancingOpenDomainTable2023} proposed a syntax and structure-aware retriever that represents each question phrase and each table header and values separately, achieving state-of-the-art results on the NQ-Tables benchmark. Likewise, \citet{liTailoringTableRetrieval2025} introduces THYME, a field-aware hybrid retriever that tailors matching strategies to different table fields (e.g., cell values vs. captions). It significantly outperforms prior baselines on open-domain table retrieval, highlighting the gains from table-specific modeling. Similarly, \citet{panEndtoEndTableQuestion2022} introduced T-RAG, an end-to-end table QA framework that combines table retrievers with sequence-to-sequence generators for retrieval-augmented generation. 

These advances, however, often rely on complex architectures or extensive fine-tuning on the TQA dataset. This raises an intriguing question: \textit{Can off-the-shelf pretrained models, when integrated into a carefully designed retrieval pipeline, achieve competitive performance to expensive domain-specialized, fine-tuned methods?} To this end, we propose \textbf{CRAFT}, a modular and scalable multi-stage retrieval framework for TQA that harnesses pretrained models without relying on dataset-specific fine-tuning.

Our contributions are as follows :
\begin{itemize}[itemsep=2pt, topsep=2pt]
  \item We design a modular, cascaded architecture that strategically combines complementary retrieval signals (lexical and semantic) to maximize recall while maintaining efficiency.
  \item Through comprehensive ablation studies, we demonstrate the robustness of CRAFT across varying retrieval thresholds \(N\), highlighting its adaptability and stability.
  \item CRAFT achieves state-of-the-art retrieval recall and end-to-end performance on NQ-Tables, and strong zero-shot performance on OTT-QA, with particularly competitive results at higher recall thresholds, demonstrating the effectiveness of training-free retrieval in real-world, open-domain settings.
  \item A mini-table centric cascaded design that reduces token footprint and embedding overhead, achieving up to $33\times$ fewer online embedding calls and $70\%$ shorter contexts. It improves retrieval and downstream QA efficiency without compromising accuracy. 
\end{itemize}

    \section{Retrieval Benchmarks}
    We used two datasets (a.) NQ-Tables~\cite{herzigOpenDomainQuestion2021}, and (b.)  OTT-QA~\cite{chen2020open} to evaluate our \textbf{CRAFT} approach. Both datasets are paired with extensive table corpora. NQ-Tables focuses on single-hop factual queries, whereas OTT-QA requires more intricate multi-hop reasoning across texts and tables. We use the subset relevant to the table for retrieval evaluation. Each query in both datasets is associated with approximately one gold table, ensuring consistency in supervision and facilitating direct comparison of retrieval performance. Table \ref{tab:dataset_stats} provides an overview of the NQ-Tables and OTT-QA benchmarks.
    
    

\begin{table}[!htbp]
\centering
\small
\setlength{\aboverulesep}{1pt}
\setlength{\belowrulesep}{1pt}
\setlength{\tabcolsep}{3.0pt}

\begin{tabular}{l|c|c|c|c}
\toprule
 & \multicolumn{2}{c|}{\textbf{NQ-Tables}} & \multicolumn{2}{c}{\textbf{OTT-QA}} \\
 & \textbf{Train} & \textbf{Test} & \textbf{Train} & \textbf{Test} \\
\midrule

\textbf{Total Query Count} & 9,594 & 966 & 41,469 & 2,214 \\
\textbf{Unique Query Count} & 9,594 & 919 & 41,469 & 2,214 \\
Avg. \# Words & 8.94 & 8.90 & 21.79 & 22.82 \\

\midrule

\textbf{Table Count} & \multicolumn{2}{c|}{169,898} & \multicolumn{2}{c}{419,183} \\
Avg. \# Rows & \multicolumn{2}{c|}{10.70} & \multicolumn{2}{c}{12.90} \\
Avg. \# Columns & \multicolumn{2}{c|}{6.10} & \multicolumn{2}{c}{4.80} \\

\midrule

\# Golden Tables / Query & 1.00 & 1.05 & 1.00 & 1.00 \\

\bottomrule
\end{tabular}

\caption{Statistics of the NQ-TABLES and OTT-QA benchmarks.}
\label{tab:dataset_stats}
\end{table}
    
    Together, these statistics underline the scalability and diversity of the evaluation setting, where NQ-Tables offer a more controlled environment and OTT-QA presents a challenging, open-domain scenario that better tests model generalization.
    \section{Our \textbf{CRAFT} Approach}

To create a retrieval setup without dataset-specific fine-tuning, we adopt a modular framework that leverages off-the-shelf models with multi-stage retrieval, with more complex models for each successive stage. As shown in Figure~\ref{fig:tablerag}, CRAFT involves pre-processing followed by stage-wise table retrieval using increasingly expressive models, culminating in answer generation via a pretrained LLM.  Our approach demonstrates that pretrained models can effectively retrieve relevant tabular data and generate accurate answers. The pipeline consists of leveraging pre-processing followed by cascaded table retrieval and end-to-end answer generation to maximize semantic alignment between input queries and tabular content using off-the-shelf embedding models.

\subsection{Preprocessing}

The preprocessing pipeline comprised three key operations designed to enhance both query and table representations for effective retrieval. Firstly, Gemini-1.5-Flash \cite{Gemini} was employed to generate sub-questions from each original query and to produce informative table titles and detailed descriptions, thereby improving query decomposition and contextual understanding. Finally, all table rows were ranked according to their semantic relevance to the queries using model (\textit{all-mpnet-base-v2}) from Sentence Transformer~\cite{reimers-2019-sentence-bert} for the NQ-Tables dataset, and Jina Embeddings v3 \cite{sturua2024jinaembeddingsv3multilingualembeddingstask} for the OTT-QA dataset. 


\subsection{Table Retrieval}

The table retrieval step follows a unified three-stage pipeline applied to NQ-Tables and OTT-QA, with dataset-specific inputs and model choices.
In the first stage, SPLADE~\cite{SPLADE2021} , a pre-trained sparse lexical-expansion retriever, was applied to a comprehensive dataset consisting of 169K  and 419K Tables in NQ-Tables and OTT-QA dataset, respectively, leveraging generated table titles, column headers, individual cell values, and table descriptions to filter and narrow down the candidate tables to the top 5,000 entries. SPLADE leverages sparse lexical expansion to efficiently score candidate passages using minimal memory, making it well-suited for handling large-scale retrieval tasks.

In the Second Stage, mini-tables are constructed by appending column headers to each of the top 5 rows and then scoring these mini-tables to produce a compact shortlist. For NQ-Table, the Sentence Transformer \textit{all-mpnet-base-v2}, a 768 dimensional pre-trained bi-encoder model optimized for semantic similarity with solid zero/few-shot retrieval performance across heterogeneous text, is used for fine-grained semantic matching due to strong sentence-level retrieval quality and robust generalization.  For OTT-QA, Jina Embeddings v3, a multilingual long-context embedding model with task adapters that improve robustness on mixed schema/numeric text, is used to better handle longer row-level contexts and numeric content typical of diverse table formats in OTT-QA. This stage selects the top K mini-tables for final reranking.

Lastly, Stage three, the top-$K$ mini-tables are re-ranked using state-of-the-art embedding models optimized for semantic precision and computational efficiency. For the NQ-Table dataset, we employ \textit{text-embedding-3-small} \cite{TEXT-EMBEDDING} and \textit{text-embedding-3-large}, for efficient semantic re-ranking. For the OTT-QA dataset, re-ranking is performed using \textit{gemini-embedding-001} \cite{lee2025gemini}, designed for long-context understanding and production-grade scalability. The Gemini embedding space enables robust semantic scoring without additional query or document augmentation, preserving the integrity of dense representations and preventing noise amplification.

Overall, this cascaded retrieval framework progressively transitions from high-recall sparse retrieval to high-precision semantic reranking, achieving efficient large-scale performance while minimizing information loss between stages. 

\subsection{End-to-End Answer Generation}

For the end-to-end table question-answering, we use the final retrieval result from CRAFT Stage 3, using three different pre-trained models: \textit{Llama3-8B} \cite{llama3}, \textit{Qwen2.5-7B} \cite{qwen2.5}, and \textit{Mistral-7B} \cite{mistral7b}. Each LLM generates answers from prompts that include a varying number of tables ($n = \{1, 3, 5, 8, 10\}$). Each prompt contains the table title, column headers, and mini-tables (top 5 rows) for $n$ tables, offering a comprehensive yet concise context to support accurate answer generation. Additionally, we experimented with the inclusion of a few-shot prompt to guide the text generation process. 
We evaluate model performance and accuracy by computing the F1 score, which quantifies the overlap between the generated answers and the reference answers, enabling a comparative analysis across different models and table settings.

    \section{Results and Analysis}
\subsection{Retrieval Performance.} 
Table \ref{tab:RetrieverResults_NQ} and Table \ref{tab:RetrieverResults_OTT} present the retrieval performance of three categories of methods Sparse, Dense, and Hybrid retrievers on the NQ-Tables and OTT-QA datasets, respectively. We conduct a comprehensive comparison across traditional sparse models such as BM25 and SPLADE, dense retrievers including TaPas~\cite{herzig2020tapas}, DPR~\cite{karpukhin-etal-2020-dense}, BIBERT ~\cite{yates-etal-2021-pretrained}, DTR, T-RAG, and SSDR, as well as hybrid approaches like Re2G, THYME, DHR~\cite{lin-etal-2023-aggretriever}, and BIBERT+SPLADE. 

\begin{table}[!htbp]
    \centering
    \small
    \setlength{\aboverulesep}{1pt}
    \setlength{\belowrulesep}{1pt}
    \setlength{\tabcolsep}{9.5pt}
    \begin{tabular}{lrcc}
        \toprule
        \textbf{Model} & \textbf{R@1} & \textbf{R@10} & \textbf{R@50} \\
        \midrule
        \multicolumn{4}{c}{\textbf{Sparse Retrieval Models}} \\
        BM25 & 18.49 & 36.94 & 52.61 \\
        SPLADE & 39.84 & 83.33 & 94.65 \\
        \addlinespace[2pt]
        \midrule
        \multicolumn{4}{c}{\textbf{Dense Retrieval Models}} \\
        BIBERT & 43.78 & 82.25 & 93.71 \\
        DPR & 45.32 & 85.84 & 95.44 \\
        TAPAS & 43.79 & 83.49 & 95.10 \\
        DTR$^{\mathsection}$ & 32.62 & 75.86 & 89.77 \\
        DTR (M) + HN$^{\dagger}$    & 47.33 & 80.96 & 91.51 \\
        DTR (S) + HN$^{\dagger}$   & 43.53 & 77.80 & 90.21 \\
        SSDR(im) & 45.47 & 84.00 & 95.05 \\
        T-RAG$^{*}$ & 46.07 & 85.40 & 95.03 \\
        \addlinespace[2pt]
        \midrule
        \multicolumn{4}{c}{\textbf{Hybrid Retrieval Models}} \\
        Re2G$^{\dagger}$ & 30.47& 66.49& 83.35\\
        BIBERT+BM25 & 35.87 & 79.63 & 94.56 \\
        DHR & 43.67 & 84.65 & 95.62 \\
        BIBERT+SPLADE & 45.62 & 86.72 & 95.62 \\
        THYME & \underline{48.55} & 86.38 & 96.08 \\
        \addlinespace[2pt]
        \midrule
        \multicolumn{4}{c}{\textbf{CRAFT Stage-wise}} \\
        Stage 1 & 34.38 & 72.90 & 91.62 \\
        \hspace{1em}+ Stage 2 & 36.65 & 82.91 & 96.08 \\
        \hspace{2em}+ Stage 3$^{*\text{s}}$ & 41.13 & \textbf{87.16} & \underline{96.84} \\
        \hspace{2em}+ Stage 3$^{*\text{l}}$ & \textbf{49.84} & \underline{86.83} & \textbf{97.17} \\
        \midrule
        \textbf{CRAFT} & \textbf{49.84} & \underline{86.83} & \textbf{97.17} \\
        \bottomrule
    \end{tabular}
    \caption{\small
    Table retrieval performance on NQ-Tables. 
    All models except \textit{CRAFT} are trained on NQ-Tables. 
    $^{*}$: results on the full test set (966 queries). 
    $^{*\text{l}}$ and $^{*\text{s}}$: use \textit{text-embedding-3-large} and \textit{text-embedding-3-small} in Stage~3, respectively. 
    $^{\mathsection}$ denotes models trained with the THYME setup. 
    $^{\dagger}$ denotes the use of the original released model checkpoints. 
    \textbf{R} = Recall. 
    Best results in \textbf{bold}; second-best are \underline{underlined}.}

    \label{tab:RetrieverResults_NQ}
     \vspace{-0.5em}
   
\end{table}

CRAFT demonstrates strong retrieval performance across both datasets, achieving state-of-the-art results on NQ-Tables and competitive zero-shot generalization on OTT-QA.

 \paragraph{Retrieval on NQ Tables.} On NQ-Tables, CRAFT progressively improves through its three-stage cascade: In Stage 1, R@10 is 72.90, and in Stage 2, it is boosted to 82.91, and Stage 3’s neural reranker further refines top-ranked results, raising R@1 to 41.13 and R@10 to 87.16. Using the higher-dimensional \textit{text-embedding-3-large} model increases R@1 to 49.84.
\paragraph{Retrieval on OTT-QA.}CRAFT maintains competitive recall at higher retrieval depths despite operating entirely without task-specific training. While fine-tuned hybrid retrievers such as THYME and BIBERT+SPLADE achieve superior Recall@1 through dataset-specific optimization, CRAFT approaches their performance at R@10 (89.88 vs. 91.10) and R@50 (96.07 vs. 96.34), with gaps of only 1.22 and 0.27 points respectively. The stage-wise improvements mirror those observed on NQ-Tables: Stage 2 delivers a substantial gain of +21.0 points at R@10 over Stage 1, while Stage 3 contributes an additional +8.0 points. This consistent progression across both datasets demonstrates that CRAFT's modular, training-free cascade effectively amplifies recall and maintains strong cross-dataset generalization, even when scaling from 169K tables (NQ-Tables) to 419K tables (OTT-QA) with more complex multi-hop reasoning requirements.



\begin{table}[!htbp]
 \vspace{-0.5em}
        \centering
        \small
        \setlength{\aboverulesep}{1pt}
        \setlength{\belowrulesep}{1pt}
        \setlength{\tabcolsep}{9.0pt}
        \begin{tabular}{lrcc}
            \toprule
            \textbf{Model} & \textbf{R@1} & \textbf{R@10} & \textbf{R@50} \\
            \midrule
            \multicolumn{4}{c}{\textbf{Sparse Retrieval Models}} \\
            BM25   & 23.98 & 51.94 & 69.11 \\
            SPLADE & 62.74 & 89.52 & 95.21 \\
            \addlinespace[2pt]
            \midrule
            \addlinespace[2pt]
            \multicolumn{4}{c}{\textbf{Dense Retrieval Models}} \\
            BIBERT & 56.82 & 86.50 & 94.26 \\
            DPR & 53.43 & 85.95 & 93.22 \\
            TAPAS & 57.86 & 86.77 & 94.04 \\
            DTR & 42.10 & 75.75 & 88.80 \\
            SSDR(im) & 56.96 & 86.22 & 93.55 \\
            \addlinespace[2pt]
            \midrule
            \addlinespace[2pt]
            \multicolumn{4}{c}{\textbf{Hybrid Retrieval Models}} \\
            BIBERT+BM25 & 59.49 & 86.81 & 94.67 \\
            DHR & 63.64 & 88.48 & 95.30 \\
            BIBERT+SPLADE & \underline{64.72} & \underline{91.01} & \textbf{96.34} \\
            THYME & \textbf{66.67} & \textbf{91.10} & \underline{96.16} \\
            \addlinespace[2pt]
            \midrule
            \addlinespace[2pt]
            \multicolumn{4}{c}{\textbf{CRAFT Stage-wise}} \\
            Stage 1 & 40.74 & 60.89 & 88.17 \\
            \hspace{1em}+ Stage 2 & 47.33 & 81.88 & 91.82 \\
            \hspace{2em}+ Stage 3 & 55.56 & 89.88 & 96.07 \\
            \midrule
            \textbf{CRAFT} & 55.56 & 89.88 & 96.07 \\
            \bottomrule
        \end{tabular}
        \caption
        {
            Table Retrieval Performance on OTT-QA. \textbf{R} denotes recall score. Best performances are shown in \textbf{bold}, and second-best are \underline{underlined}.
            All models, except \textit{CRAFT}, are trained on OTT-QA. SPLADE in the table is the supervised baseline, whereas CRAFT Stage 1 uses the base SPLADE model.
        }
        \label{tab:RetrieverResults_OTT}
        \vspace{-1em}
    \end{table}

\subsection{Robustness of \textbf{CRAFT} Retrieval.}
    
To assess robustness under query paraphrasing, we generated perturbed versions of each question using \textit{Gemini~2.5~Flash}, while keeping the ground-truth answers unchanged. For example: 
    
    \begin{framed}
    \vspace{-0.25em}
    {
    \small
    \begin{description}
     \vspace{-1.0em}
     
        \item[Original]: Philadelphia is known as the city of what?
        \vspace{-0.2em}
        \item[Perturbed]: What descriptor follows ``the city of'' when referencing Philadelphia?
         \vspace{-0.2em}
        \item[Answer]: Brotherly Love
         \vspace{-2.0em}
    \end{description}
    }
    \end{framed}
    
\noindent The prompt used for generating the perturbed queries is detailed in Figure~\ref{fig:prompt_query}.

\noindent \paragraph{Analysis.} Table \ref{tab:perturbed_dtr_tabret_delta} reports recall scores on perturbed queries for DTR, trained on NQ-Tables with and without hard negatives, and our three-stage CRAFT pipeline. We chose it as a baseline because it is a traditional, well-known, and widely adopted model. The checkpoint used in our experiments is publicly available and has been finetuned on the NQ-Tables dataset, making it an ideal baseline for comparing our Cascade Approach against a standard dense embedding method. Additionally, it serves as a common baseline used in many papers that report scores on the NQ-Tables Dataset, like LI-RAGE, DTR, THYME, CLTR, SSDR, and T-RAG, facilitating easier comparison.

\begin{table}[htbp]
        \small
        \centering
    \setlength{\aboverulesep}{1pt}
    \setlength{\belowrulesep}{1pt}
    \setlength{\tabcolsep}{5.5pt}
        \begin{tabular}{lcccc}
            \toprule
            \textbf{Model} & \textbf{R@1} & \textbf{R@10} & \textbf{R@50} & \textbf{$\Delta$ (Avg.)} \\
            \midrule
            \multicolumn{5}{c}{\textbf{Original Query}} \\[3pt]
            DTR (M)                 & 38.74 & 75.73 & 88.36 & - \\
            DTR (M) + HN            & \textbf{47.33} & 80.96 & 91.51 & - \\
            DTR (S)                 & 39.28 & 73.88 & 87.60 & - \\
            DTR (S) + HN            & 43.53 & 77.80 & 90.21 & - \\
            \bf CRAFT                  & 41.13 & \textbf{87.16} & \textbf{96.84} & - \\
            \midrule
            \multicolumn{5}{c}{\textbf{Perturbed Query}} \\ [3pt]
            DTR (M)                 & 27.09 & 66.27 & 84.33 & - 8.38 \\
            DTR (M) + HN            & 37.87 & 76.06 & 88.47 & - 5.80 \\
            DTR (S)                 & 25.25 & 61.81 & 78.24 & - 11.82 \\
            DTR (S) + HN            & 35.04 & 70.73 & 85.85 & - 6.64 \\
            \midrule
            \multicolumn{5}{c}{\textbf{CRAFT Stage-wise}} \\[3pt]
            Stage 1                 & 33.40 & 68.22 & 87.05 & - 3.41\\
            \hspace{1em}+ Stage 2   & 34.05 & 73.18 & 88.92 & - 6.5\\
            
            \hspace{2em}+ Stage 3 $^{*\text{s}}$  & \textbf{41.02} & \textbf{86.83} & \textbf{96.08} & - 0.04 \\
            \bottomrule
        \end{tabular}%
        \caption{\small Performance of DTR vs CRAFT with perturbed queries on NQ-Tables. \textbf{M} and \textbf{S} denote Medium and Small model sizes respectively. $\Delta$ is the average change in Recall relative to the original query scores, positive values indicate improvement, and negative values indicate degradation. \textbf{HN} represents a model trained with Hard Negatives.
        $^{*\text{s}}$ indicate the use of \textit{text-embedding-3-small} model in Stage~3.}
        \label{tab:perturbed_dtr_tabret_delta}
    \end{table}

With the original queries, \textbf{CRAFT} already outperforms the DTR baselines. When using the perturbed queries, both DTR Medium and Small suffer substantial average drops of –8.38 and –11.82 points, respectively. The DTR version trained with hard negatives reduces but does not eliminate this degradation (avg = –5.80 for DTR (M) and –6.64 for DTR (S)). By contrast, CRAFT’s full three-stage pipeline nearly fully recovers its original performance with a slight decrease in paraphrasing.

    \paragraph{Stage-wise Analysis.} Analysis over stages further shows incremental gains: Stage 1  drops performance under perturbation by 3.4 \%, Stage 2 partly recovers it, and Stage 3 restores the original recall, as shown in Table \ref{tab:perturbed_dtr_tabret_delta}. These results highlight CRAFT’s superior robustness to query rephrasing compared to DTR.

\subsection{End-to-End Results.}
    We evaluate the end-to-end performance of CRAFT across multiple pretrained language models and retrieval configurations. For end-to-end QA, our approach aligns with recent work, such as THYME and EE-BIBERT, which report results on larger open-source models like Llama3-8B, Qwen2.5-7B, and Mistral-7B and closed-source model GPT-4o. 
    \paragraph{Results on NQ-Tables.} As shown in Table~\ref{tab:End-to-End Results}, CRAFT consistently outperforms several state-of-the-art baselines across different models, for varying numbers of retrieved tables ($n = {1, 3, 5}$). These results highlight the robustness of our modular, training-free approach. 
    Beyond higher recall, mini-table filtering and compact context windows reduce irrelevant content and allow more evidence(more retrieved tables) per query, so even semantically correct non-gold tables can still support higher end-to-end F1.
    \begin{table*}[!htbp]
             \small
            \centering
    \setlength{\aboverulesep}{0pt}
    \setlength{\belowrulesep}{1.5pt}
    \setlength{\tabcolsep}{5.5pt}
                \begin{tabular}{l|ccc|ccc|ccc}
                    \toprule
                    \textbf{Retriever} & \multicolumn{3}{c|}{\textbf{n=1}} & \multicolumn{3}{c|}{\textbf{n=3}} & \multicolumn{3}{c}{\textbf{n=5}} \\
                     & \textbf{Mistral} & \textbf{Llama3} & \textbf{Qwen} & \textbf{Mistral} & \textbf{Llama3} & \textbf{Qwen2.5} & \textbf{Mistral} & \textbf{Llama3} & \textbf{Qwen2.5} \\
                    \midrule
                    BIBERT & 32.93 & 32.66 & 34.80 & 34.6 & 33.16 & 37.92 & 35.30 & 34.28 & 37.01 \\
                    SPLADE & 29.61 & 32.07 & 31.90 & 35.42 & 37.17 & 35.79 & 34.95 & 33.88 & 37.35 \\
                    SSDR\_in & 32.76 & 34.25 & 33.42 & 36.95 & 38.14 & 37.14 & 36.60 & 37.89 & 33.92 \\
                    BIBERT-SPLADE & 32.67 & 32.66 & 33.24 & 33.59 & 33.66 & 36.76 & 35.71 & 33.92 & 37.02 \\
                    THYME & 35.48 & 36.14 & 37.28 & 37.59 & 39.16 & 40.28 & 37.20 & 39.29 & 41.20 \\
                    \midrule
                    \textbf{CRAFT} & \textbf{39.13} & \textbf{38.31} & \textbf{39.73} & \textbf{45.28} & \textbf{40.76} & \textbf{43.52} & \textbf{44.53} & \textbf{40.55} & \textbf{46.49} \\
                    \bottomrule
                \end{tabular}
            \vspace{-0.75em}
            \caption{\small Performance comparison (F1) across different retrievers, LLMs, and values of $n$ with NQ-Tables. The evaluated models include \textit{Mistral-7B-Instruct}, \textit{Llama3-8B-Instruct}, and \textit{Qwen2.5-7B-Instruct}, corresponding to \textbf{Mistral}, \textbf{Llama3}, and \textbf{Qwen}, respectively. \textbf{Bolded} values indicate top results.}
            \label{tab:End-to-End Results}
        \end{table*}
    
 \begin{table}[!ht]
        \small
        \centering
    \setlength{\aboverulesep}{2pt}
    \setlength{\belowrulesep}{2pt}
    \setlength{\tabcolsep}{5.5pt}
        
        \begin{tabular}{l|cc}
            \toprule
            \textbf{Model} & \textbf{n =1} & \textbf{n=5} \\
            \midrule
                RAG \cite{lewisRetrievalaugmentedGenerationKnowledgeintensive2020a}                         &     -   & 39.67 \\
                DPR-RAGE \cite{linLIRAGELateInteraction2023}                    &     -   & 49.68 \\
                LI-RAGE \cite{linLIRAGELateInteraction2023}                    &     -   & 54.17 \\
            \midrule
                \textbf{CRAFT}\\
                    \quad- Qwen2.5-72B-Instruct       & 48.75 & 49.17 \\
                    \quad- Llama-3.3-70B-Instruct     & 46.21 & 55.16 \\
                    \quad- Llama-3.1-70B-Instruct     & 49.75 & 56.94 \\
                    \quad- Mistral-Small-Instruct     & \underline{50.26} & \underline{57.14}\\
                    \quad- GPT-4o     & \textbf{56.76} & \textbf{67.72}\\
            \bottomrule
        \end{tabular}
        \caption{\small F1 scores of \textbf{CRAFT} on NQ-Tables with larger open source and closed source LLMs compared to dataset-specific baselines for $n=1$ and $n=5$ retrieved tables. Bold indicates the best performance.}
    \label{tab:f1_topk_summary}
    \end{table}
    \paragraph{Effect of Larger LLMs.} Table~\ref{tab:f1_topk_summary} further emphasizes the effectiveness of CRAFT, particularly when employing larger, instruction-tuned LLMs. Compared to prior dataset-specific baselines, CRAFT consistently delivers superior F1 scores. We find that using just a single retrieved table ($n=1$), \textit{GPT 4o} achieves an F1 score of $56.76$, already surpassing the best-performing baseline, LI-RAGE ($54.17$), even though it benefits from deeper retrieval ($n=5$). With increased retrieval depth ($n=5$), \textit{GPT-4o} further improves to $67.72$, highlighting significant gains from additional context. Similarly, \textit{Mistral-Small-Instruct} and \textit{Llama-3.1-70B-Instruct} achieve notable scores of $57.14$ and $56.94$ at $n=5$, clearly outperforming LI-RAGE (current SOTA). 

    \paragraph{Effect of Few-Shots.}
        To evaluate the scalability and robustness of CRAFT, under zero-shot and few-shot prompting as the number of retrieved tables increases ($n = 1, 3, 5, 8, 10$). As shown in Table~\ref{fig:f1_extended} (Appendix Section~\ref{appdx:End-to-End}), all models exhibit a clear upward trend in F1 score with deeper retrieval, demonstrating that structured multi-stage retrieval alone substantially narrows the gap with dataset-specific fine-tuning. However, incorporating a small number of in-context examples does not consistently yield significant improvements across models. Moreover, performance gains taper beyond $n=5$, suggesting an optimal retrieval depth that balances information richness with the risk of noise accumulation. This effect becomes more pronounced as both $n$ and the number of few-shot examples increase, leading to greater performance variability.
    \paragraph{Result with OTT-QA.} We further provide end-to-end performance on the OTT-QA dataset under an oracle setting, where retrieved tables are paired with the gold passage. Under this oracle setup, where retrieved tables are paired with their corresponding gold passages to isolate the effect of retrieval quality, our end-to-end OTT-QA evaluation shows that Stage~3 (MT) consistently achieves the best balance between recall and precision. As this setup is not directly comparable to prior baselines, we present these results in the Appendix Section~\ref{appdx:end2end-ottqa} for completeness.
    

These results underscore that \textbf{CRAFT}'s structured, training-free retrieval methodology not only outperforms established baselines but also scales effectively with increased retrieval depth and few-shot prompting. This highlights the strength of CRAFT's modular design in enabling adaptable, efficient, and high-performing Table-QA without requiring any dataset-specific fine-tuning.

\subsection{Computational Efficiency}

     \begin{figure*}[!htbp]
    \centering
        \includegraphics[width=1.0\linewidth]{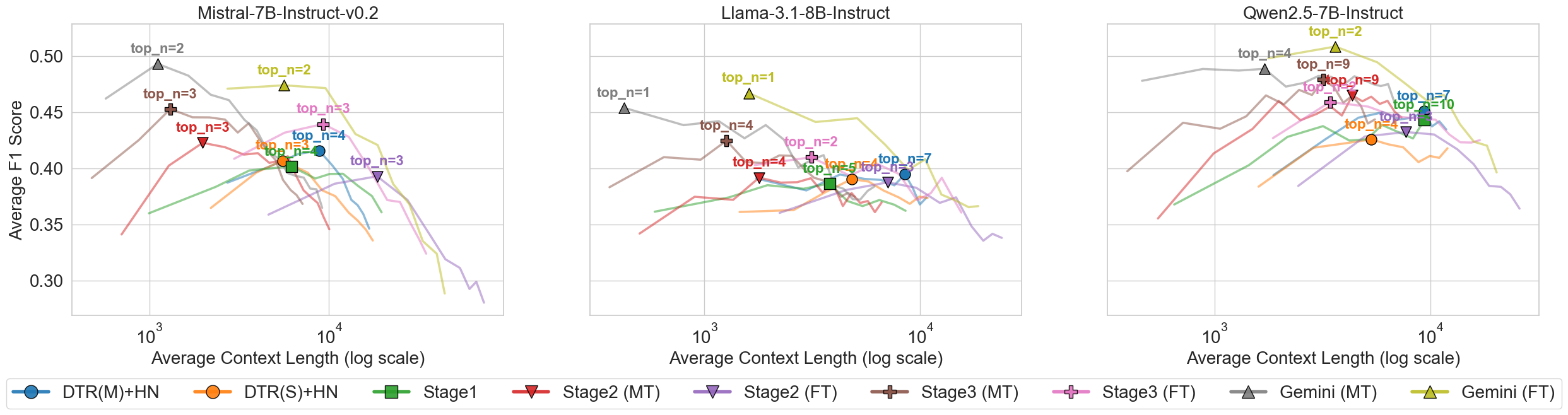}
              \caption{\small End-to-end performance comparison across context lengths with the NQ dataset. The figure illustrates the relationship between average context length and F1 score across three models, evaluated over ten different numbers of retrieved tables ($n$) on the NQ dataset. FT denotes the use of the Full Table for end-to-end evaluation, while MT represents the Mini-Table variant.
              }
              \label{fig:context_f1_score}
     \end{figure*} 

        Unlike end-to-end trained baselines that shift most computation to offline training and then rely on a single, fully embedded corpus at inference time, CRAFT's cascaded architecture strategically concentrates expensive computation only where it provides the greatest benefit. This modular design offers two key advantages: (1) it avoids costly retraining when adapting to new domains, and (2) it enables seamless integration of emerging embedding models by simply swapping components without architectural changes. 
        \paragraph{Retrieval Efficiency.}  CRAFT shifts the bulk of computation to an offline pre-computation phase, drastically reducing online embedding calls. We build a SPLADE inverted index over all tables using sparse, lightweight lexical representations and pre-encode each table row with the Stage 2 dense ranker model exactly once. At inference, CRAFT executes only three query embeddings plus 5,100 mini-table embeddings per query:
    
    \begin{itemize}[itemsep=1pt,topsep=1pt]
        \item Stage 1 (Sparse Retrieval): One query encoding + inverted-index lookup over 169,898 (NQ-Tables) / 419,183 (OTT-QA) tables.
        \item Stage 2 (Dense Pruning): One \textit{all-mpnet-base-v2} query embedding + sublinear-time ANN search over all precomputed row embeddings to select the top 5,000 mini-tables, avoiding the $\mathcal{O}(N)$ cost of brute-force search.
        \item {Stage 3 (Re-ranking): One \textit{text-embedding-3-small} query embedding + embeddings for the top 100 mini-tables.}
    \end{itemize}
    
    By contrast, a fully dense baseline requires 169,898 table embedding calls. Stage 1 pruning reduces the candidate set by over 97\% while retaining 99.56\% recall Table~\ref{tab:RetrieverResultsStageWise} cutting embedding calls by over 33$\times$ without sacrificing accuracy.
\paragraph{Computational Trade-offs .} Stage 2 employs a lightweight dense encoder to produce embeddings for the generated mini-tables on-the-fly during inference. While this increases per-query latency compared to retrieving from a pre-embedded corpus, it substantially reduces irrelevant content, thereby saving valuable context tokens for downstream processing. This design manages the trade-off between per-query cost and end-to-end performance by filtering noise early in the pipeline, which significantly benefits Stage 3 retrieval and LLM-based QA, both highly sensitive to token budgets. CRAFT’s modular design allows seamless integration of new embedding models, as efficiency is primarily governed by token budgets and per-query embedding calls.

\paragraph{End-to-End Efficiency.} While cascading retrieval improves performance with reduced offline computational overhead, computational efficiency also remains critical for the end-to-end QA task.

    \begin{table}[!htbp]
    \small
    \centering
    \setlength{\aboverulesep}{1pt}
    \setlength{\belowrulesep}{1pt}
    \setlength{\tabcolsep}{5.0pt}
        \begin{tabular}{l|c|r|r}
            \toprule
            \textbf{Model} & \textbf{n} & \textbf{Sub-Table} & \textbf{Full Table} \\
            \midrule
            \multirow{3}{*}{Llama-3.1-8B-Instruct} & 1 & 362.70 & 1782.80 \\
             & 3 & 958.79 & 4841.75 \\
             & 5 & 1561.34 & 8231.99 \\
             & 8 & 2544.36 & 12831.89 \\
             & 10 & 3213.31 & 15798.45 \\
            \midrule
            \multirow{3}{*}{Mistral-7B-Instruct-v0.2} & 1 & 476.58 & 3314.03 \\
             & 3 & 1305.60 & 10183.81 \\
             & 5 & 2158.50 & 18252.36 \\
             & 8 & 3566.04 & 29321.20 \\
             & 10 & 4537.62 & 36191.87 \\
            \midrule
            \multirow{3}{*}{Qwen2.5-7B-Instruct}  & 1 & 393.68 & 1951.08 \\
            & 3 & 1060.45 & 5319.05 \\
            & 5 & 1731.25 & 9009.43 \\
            & 8 & 2824.83 & 14046.00 \\
            & 10 & 3568.37 & 17292.88 \\
            \bottomrule
        \end{tabular}%
        \vspace{-6pt}
        \caption{Comparison of average token counts between sub-table and full-table contexts across models and top-$k$ values.}
        \label{tab:token_model_comparison}
        \vspace{-14pt}

    \end{table}
\paragraph{Effect of Mini-Tables.} Table~\ref{tab:token_model_comparison} compares the average token counts for each model when using only the top-$k$ sub-table context versus concatenating the entire table. Across all three models and $k \in {1, 3, 5}$, sub-table inputs (CRAFT mini-tables) reduce token length by over 70\%. For example, Llama3 drops from 1782.80 tokens at $k=1$ to 362.70; Mistral from 3,314.03 to 476.58; and Qwen2.5 from 1,951.08 to 393.68. This substantial reduction directly lowers inference costs and improves runtime efficiency. By selectively presenting only the most relevant table rows fragments, \textbf{CRAFT} retains essential semantic content while avoiding the diminishing returns and latency overhead associated with full-table encoding. 
\paragraph{Effect of Token Reduction.} Figure~\ref{fig:context_f1_score} compares the end-to-end performance of three instruction-tuned models across varying context lengths on the NQ dataset. The curves reveal a clear trade-off between context breadth and noise accumulation: smaller retrieved sets (low n) often underperform because they may exclude the gold table, while larger retrieved sets (high n) increase the risk of including irrelevant content that dilutes attention and lowers F1. This pattern is particularly evident for Qwen2.5-7B-Instruct, where performance peaks at top\_n = 9 using the Mini-Table (MT) configuration, showing that our multi-stage retrieval strategy effectively identifies an optimal context length that balances relevance and compactness. Stage 3 (MT) consistently achieves superior results at shorter context lengths, highlighting the benefit of reducing redundant information while retaining sufficient evidence for reasoning. 

\section{Ablation Study}\label{sec:ablation_results}
As Table \ref{tab:RetrieverResultsStageWise} shows, our three-stage cascade smoothly shifts from broad coverage to pinpoint accuracy. In Stage 1, the SPLADE filter finds nearly all relevant tables at deeper ranks (91.62 \% at R@50) but only 34.38 \% at the very top. However, SPLADE achieves a R@5000 of 99.59\%, indicating performance saturation and leaving minimal scope for further enhancement. Stage 2’s Sentence-Transformer reranker then delivers most of the mid-level gains, raising R@10 by over 10 points and boosting R@50 above 96\%, while still keeping high deep recall. The top 5,000 tables returned by Stage 1 are passed to Stage 2, which performs dense semantic reranking over these candidates. In Stage 2, we achieve the highest recall of 97.82$\%$, which is a significant improvement over the Stage 1 retrieval. These 100 tables are carried forward to Stage 3, our final retrieval step. Finally, Stage 3’s neural reranker sharpens the top results, lifting R@1 by 4.5 points to 41.13 \% and increasing R@10 to 87.16 \%. Each stage builds upon the last, improving accuracy step by step without requiring any additional training.

\begin{table*}[!htbp]
    \small
    \centering
    \setlength{\aboverulesep}{0pt}
    \setlength{\belowrulesep}{1pt}
    \setlength{\tabcolsep}{4.0pt}
    \begin{tabular}{l|cccccc|cccccc}
        \toprule
        & \multicolumn{6}{c|}{\textbf{NQ-Tables}} & \multicolumn{6}{c}{\textbf{OTT-QA}} \\
        \textbf{Model} & \textbf{R@1} & \textbf{R@10} & \textbf{R@50} & \textbf{R@100} & \textbf{R@500} & \textbf{R@5K} & 
        \textbf{R@1} & \textbf{R@10} & \textbf{R@50} & \textbf{R@100} & \textbf{R@500} & \textbf{R@5K} \\
        \midrule
        Stage 1                 & 34.38 & 72.90 & 91.62 & 94.89 & 98.91 & \bf \fbox{99.56} & 40.74 & 60.89 & 88.17 & 91.64 &96.93 & \bf \fbox{99.28} \\
        Stage 2   & 36.65 & 82.91 & 96.08 & \bf \fbox{97.82} & \textbf{98.91} & - & 47.33 & 87.35 & 91.82 &94.21 & \bf \fbox{97.38} & - \\
        Stage 3   & 41.13 & 87.16 & 96.84 & -  & - & - & 55.56 & 89.88 & 96.07  &- & - & - \\
        \midrule
        \textbf{CRAFT} &\bf 41.13 & \textbf{87.16} &\textbf{96.84}& - & - & - & \textbf{55.56} & \textbf{89.88} & \textbf{96.07}   &  \\
        \bottomrule
    \end{tabular}%
\caption{\small Table Retrieval Performance on NQ-Tables and OTT-QA datasets with various stages of CRAFT. R denotes Recall Score. Best performances are in \textbf{Bold}. \fbox{Boxed values} indicate the number of tables passed to the next stage for re-ranking. }
\label{tab:RetrieverResultsStageWise}
\end{table*}

\paragraph{Error Analysis} To further understand the performance gaps, we provide a qualitative error analysis in Appendix~\ref{appdx:error_analysis}. We find that most failure cases arise from near-duplicate tables with highly similar schemas or ambiguous entity mentions, which are difficult to disambiguate without task-specific supervision. Additionally, even when the gold table is retrieved, using full tables can introduce noise that hinders answer extraction, whereas our mini-table representation isolates relevant rows and enables the LLM to generate correct answers more reliably.

\label{sec:results_ablations_results}
\paragraph{Stage 1: Sparse Retrieval with Context Enrichment} We test the effect of table descriptions and query decomposition using BM25 and SPLADE. The results for Stage 1 are shown in Table~\ref{tab:Stage1TableDescription} and Table~\ref{tab:recall_results}. Both retrievers exhibit improvement when employing Question Decomposition. Specifically, SPLADE achieves gains of 2.18\% and 4.03\% at R@10 and R@50, respectively. With the table description, SPLADE achieved a consistent increase in recall. Notably, SPLADE achieves an R@5000 of 99.59\%, indicating performance saturation and leaving minimal scope for further enhancement. In contrast, BM25 demonstrates substantial improvement with the inclusion of table descriptions, clearly benefiting from context-enriched lexical matching.

\begin{table}[H]
    \centering
    \small
\setlength{\aboverulesep}{2pt}
\setlength{\belowrulesep}{2pt}
\setlength{\tabcolsep}{4pt}
        \begin{tabular}{l|ccccc}
            \toprule
            \textbf{Table} & \textbf{R@1} & \textbf{R@10} & \textbf{R@50} & \textbf{R@1k} & \textbf{R@5k} \\
            \midrule
            \multicolumn{6}{c}{\textbf{BM25}} \\
            w/o Desc & 17.95 & 34.38 & 47.44 & 70.73 & 82.48 \\
            + Desc & 22.74 & 45.16 & 62.57 & 86.61 & 93.14 \\
            \midrule
            \multicolumn{6}{c}{\textbf{SPLADE}} \\
            w/o Desc & 31.06 & 69.05 & 86.85 & 98.45 & 99.59 \\
            + Desc & \textbf{32.64} & \textbf{70.62} & \textbf{87.59} & \textbf{98.91} & \textbf{99.67} \\
            \bottomrule
        \end{tabular}%
    \caption{\textbf{Stage 1:} Effect of Table Descriptions on Retrieval Performance over NQ-Tables. \textbf{R} denotes Recall Score. \textbf{N}=169,898 Tables in Corpus.}
    \label{tab:Stage1TableDescription}
\end{table}
\begin{table}[H]
\small
    \centering
        \setlength{\aboverulesep}{2pt}
\setlength{\belowrulesep}{2pt}
\setlength{\tabcolsep}{4pt}
    {%
        \begin{tabular}{l|ccccc}
            \toprule
            \textbf{Question} & \textbf{R@1} & \textbf{R@10} & \textbf{R@50} & \textbf{R@1k} & \textbf{R@5k} \\
            \midrule
            \multicolumn{6}{c}{\textbf{BM25}} \\
            Original & 22.74 & 45.16 & 62.57 & 86.61 & 93.14 \\
              + SubQues & 24.15 & 53.54 & 71.38 & 91.08 & \textbf{96.84} \\
            \midrule
            \multicolumn{6}{c}{\textbf{SPLADE}} \\
            Original & 32.64 & 70.62 & 87.59 & 98.91 & \textbf{99.67} \\
              + SubQues & \textbf{34.38} & \textbf{72.90} & \textbf{91.62} & \textbf{98.91} & 99.56 \\
            \bottomrule
        \end{tabular}%
    }
    \caption{\textbf{Stage 1:} Effect of Query Decomposition- Retrieval Performance on NQ-Tables. \textbf{R} denotes Recall Score. \textbf{N}=169,898 Tables in Corpus.} 
    \label{tab:recall_results}
\end{table}
\paragraph{Stage 2: Dense Semantic Reranking and Irrelevant Context Reduction} We apply the Stage 2 dense model to rerank the top 5000 tables. Since dense embedding models have limited context length, we rank the rows of each table and retain the 5 most relevant rows per table. Table \ref{tab:TableRetrievalStage2} shows recall gains from filtering irrelevant rows, with R@10 rising from 79.21 to 82.91 and R@50 from 94.66 to 96.08. Dense scoring better captures the semantic relevance missed in Stage 1. 

\begin{table}[!htbp]
    \small
    \centering
\setlength{\aboverulesep}{2pt}
\setlength{\belowrulesep}{2pt}
\setlength{\tabcolsep}{2pt}
    \begin{tabular}{l|ccccc}
            \toprule
            \textbf{Table Text} & \textbf{R@1} & \textbf{R@10} & \textbf{R@50} & \textbf{R@100} & \textbf{R@1k} \\
            \midrule
            Original* & \textbf{37.64 }& 79.76 & 94.45 & 97.82 & 99.45 \\
            \hspace{0.5em}--Description  & 34.82 & 79.21 & 94.66 & 97.39 & \textbf{99.45} \\
            \hspace{1em}--Irrelevant Rows& 36.65 & \textbf{82.91} & \textbf{96.08} & \textbf{97.82} & 98.91 \\
            \bottomrule
        \end{tabular}%
    \caption{\textbf{Stage 2:} Effect of Irrelevant Context Reduction- Retrieval using NQ-Tables. Original* is the title, header, all cell values, and description. \textbf{N2}=5000 Tables in Corpus. Best performances are in \textbf{Bold}.}
    
    \label{tab:TableRetrievalStage2}
\end{table}
\paragraph{Stage 3: Neural Reranking}

Finally, we rerank the top 300 and 100 mini-tables. Table \ref{tab:stage3_embedding_results} shows the reranking results. Even without fine-tuning, neural reranking significantly sharpens final table rankings.
\begin{table}[!htbp]
        \centering
        \small
    \setlength{\aboverulesep}{2pt}
    \setlength{\belowrulesep}{2pt}
    \setlength{\tabcolsep}{10.0pt}
            \begin{tabular}{l|ccc}
                \toprule
                \textbf{N3} & \textbf{R@1} & \textbf{R@10} & \textbf{R@50} \\
                \midrule
                300 & 39.03 & 86.02 & \textbf{96.89} \\
                100 & \textbf{41.13} & \textbf{87.16} & 96.84 \\
                \bottomrule
            \end{tabular}
        \caption{\textbf{Stage 3:} Effect of Neural Reranking: Retrieval Performance using \textit{text-embedding-3-small} model on NQ-Tables. \textbf{N3} is the number of tables in the corpus. Best performances are in \textbf{Bold}.}
        \label{tab:stage3_embedding_results}
    \end{table}

    These findings confirm the effectiveness of the cascaded design, which begins with broad lexical and semantic coverage, progressively refines relevance through dense pruning, and culminates in lightweight neural reranking for precise top-rank. Collectively, they highlight how each stage contributes a distinct function within the pipeline, yielding an accurate and computationally efficient retrieval framework.

    \section{Conclusion}

    We present CRAFT, a modular, training-free cascaded retrieval framework for Table Question Answering that achieves competitive performance compared to state-of-the-art systems that require dataset-specific fine-tuning. CRAFT achieves new state-of-the-art retrieval Recall@1 on NQ-Tables, maintains high robustness against query paraphrasing, and delivers strong end-to-end QA performance when paired with modern LLMs. By decoupling retrieval from domain-specific fine-tuning, CRAFT offers a scalable and adaptable paradigm for dynamic tabular domains where rapid deployment and data scarcity are critical issues. Future work will extend CRAFT's capabilities to enhance its reasoning through a fully integrated, multi-hop retrieval system to tackle complex benchmarks, such as OTT-QA, without relying on oracle passages. 
    These directions will further advance CRAFT toward a unified, efficient, and generalizable solution for open-domain table understanding.

\section{Limitations}

    While this study demonstrates the effectiveness of a training-free, cascaded approach, its scope has certain boundaries. The evaluation centers on single-hop question answering over NQ-Tables and OTT-QA, leaving multi-hop reasoning tasks that demand synthesis across multiple tables and passages unexplored. One promising extension adds a lightweight contrastive objective after Stage 3 to sharpen R@1 by separating structurally similar yet semantically distinct tables, which we leave for future work.
    
    The end-to-end QA evaluation for OTT-QA ran in an oracle setting, using gold passages to isolate retrieval performance, falling short of a full open-retrieval pipeline. Our approach targets text-based table retrieval and does not handle multimodal tables containing images or other non-textual data. Finally, although CRAFT's offline pre-computation runs efficiently, the pipeline's multi-stage design pushes per-query processing time upward.

    \section{Acknowledgements}
We thank the anonymous reviewers and the Area Chair for their constructive feedback. We are grateful to the members of the Complex Data Analysis and Reasoning Lab for their helpful discussions and contributions to the project’s ideation. We also thank Aman Singh and Prasham Titiya for their insightful input during the early stages of this work. We also acknowledge the support of the USA National Science Foundation under grant \#2047488 and the Rensselaer-IBM Future of Computing Research Collaboration (FCRC).

\section{Ethics Statement}

This research adheres to standard ethical practices in NLP. We built our approach using publicly available datasets, NQ-Tables and OTT-QA, that contain no sensitive or personally identifiable information. For generating table titles, descriptions, and query expansions, we leveraged \textit{Gemini 1.5 Flash} alongside other embedding models and language models, all used within their specified usage guidelines. Since our framework is training-free, we didn't need to collect or annotate new data. That said, we recognize that our method can reflect biases present in the underlying datasets and pre-trained models. To address this, we kept our experiments in controlled settings and report results with full transparency. We've included enough implementation details to make our work reproducible, and any AI-assisted writing tools were used solely to refine clarity and all outputs were carefully reviewed and verified by the authors.

    \bibliography{bib}
    \appendix
    \section{Hyperparameter Configuration}

    \begin{table}[!htbp]
    \centering
    \small
    \begin{tabular}{l|c}
        \toprule
        \textbf{Configuration Parameter} & \textbf{Value} \\
        \midrule
        Batch Size & 144 \\
        Learning Rate & $1 \times 10^{-5}$ \\
        Training Epochs & 50 \\
        Optimizer & Adam \\
        Initialization & BERT-base \\
        THYME Dropout ($p_{sem}, p_{lex}$) & 0.15 \\
        THYME FLOPS Reg. ($\lambda_q, \lambda_t$) & $1 \times 10^{-4}$ \\
        \bottomrule
    \end{tabular}
    \caption{Training and inference configuration used for all baselines.}
    \label{tab:baseline_config}
\end{table}

    Table \ref{tab:baseline_config} outlines the training and inference configuration adopted for baseline retrievers. These configurations were chosen to ensure stable convergence across both datasets while maintaining generalization in retrieval performance. The combination of moderate dropout, lightweight regularization, and a conservative learning rate enabled consistent optimization across heterogeneous data distributions, striking a balance between efficiency and representational expressiveness. For Re2G , we directly used the publicly released checkpoints fine-tuned on NQ Dataset.

    \section{Prompt Used For Experiments}
\begin{figure}[!htbp]
        \centering
                  \frame{\includegraphics[width=\columnwidth]{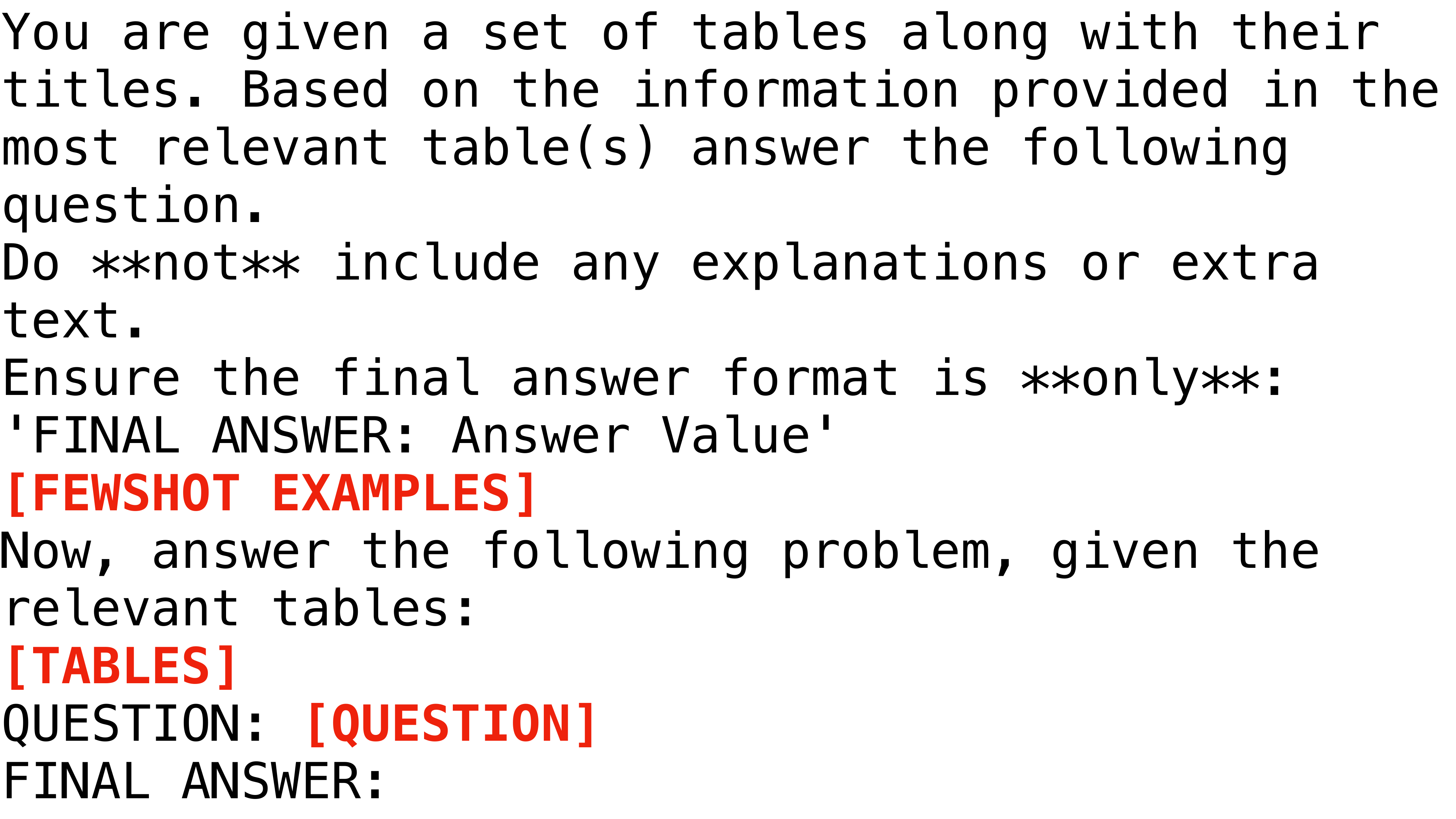}}
                  \caption{Prompt used for End-to-End question answering.}
                  \label{fig:prompt_e2e}
    \end{figure}

\begin{figure*}[!htbp]
        \centering
                  \frame{\includegraphics[width=0.98\linewidth]{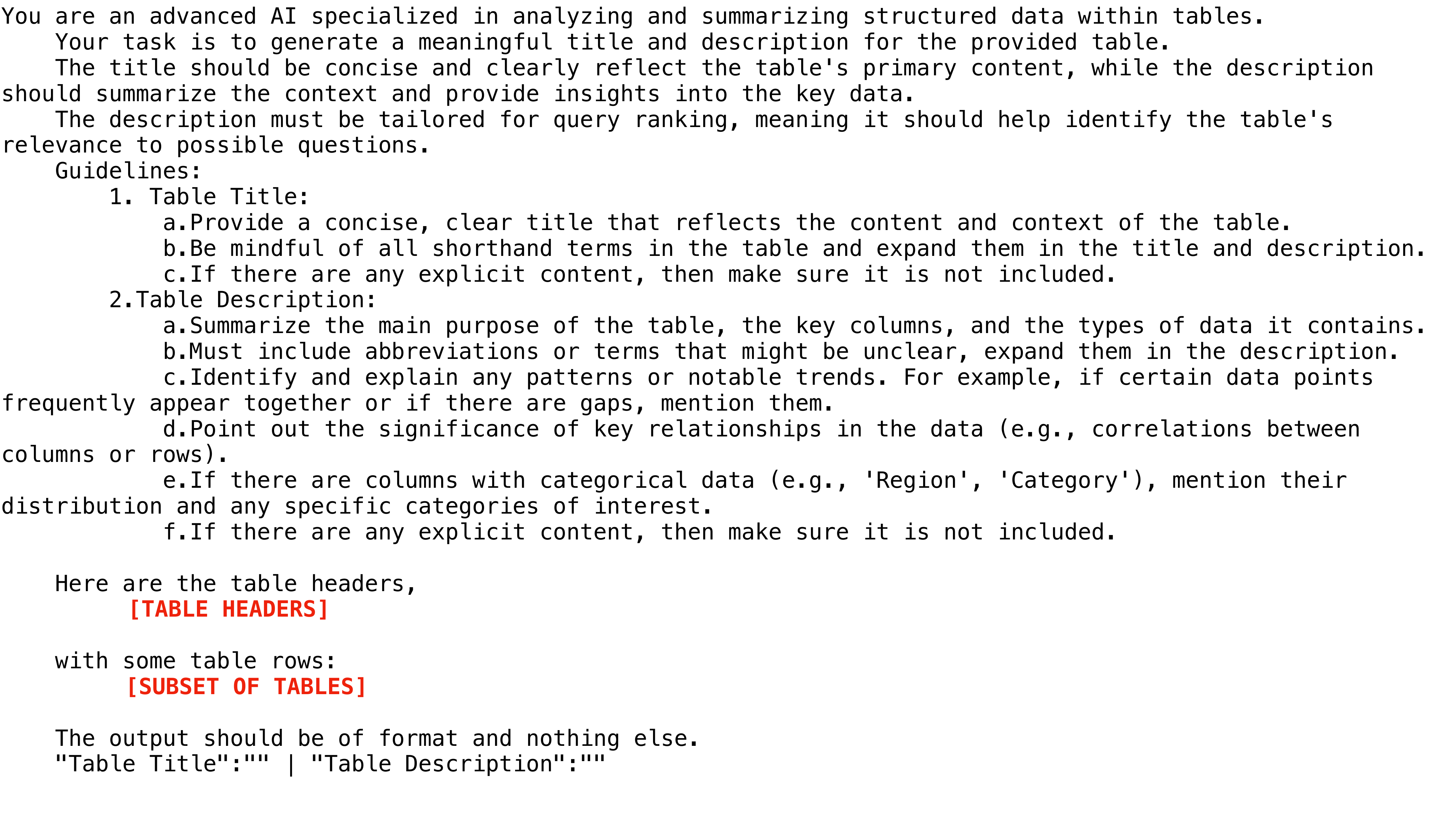}}
                  \caption{Prompt used for Table Decomposition.}
                  \label{fig:prompt_table}
        \end{figure*}

\begin{figure*}[!htbp]
        \centering
                  \frame{\includegraphics[width=0.98\linewidth]{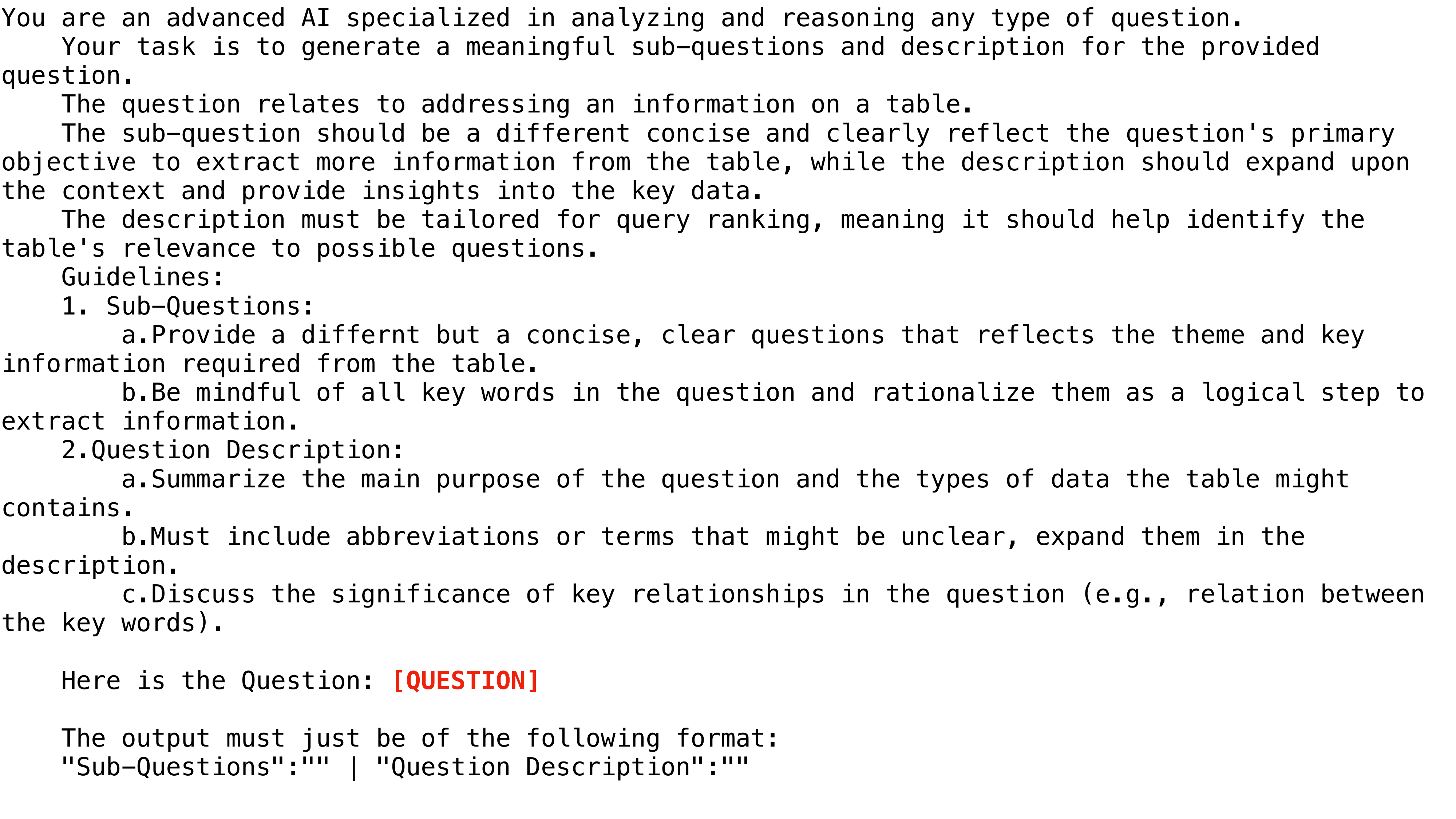}}
                  \caption{Prompt used for Query Expansion.}
                  \label{fig:prompt_query}
        \end{figure*}

    The prompt used for end-to-end question answering in Figure~\ref{fig:prompt_e2e}, the table description prompt in Figure~\ref{fig:prompt_table} and the question decomposition prompt is shown in Figure~\ref{fig:prompt_query}.


    \section{Further End-to-End Analysis}\label{appdx:End-to-End}
        \begin{table*}[htbp]
        \centering
        \renewcommand{\arraystretch}{1.2}
        \resizebox{\textwidth}{!}{%
        \begin{tabular}{l|ccc|ccc|ccc|ccc|ccc}
        \toprule
        \textbf{Retriever} & \multicolumn{3}{c|}{\textbf{n=1}} & \multicolumn{3}{c|}{\textbf{n=3}} & \multicolumn{3}{c}{\textbf{n=5}} & \multicolumn{3}{c}{\textbf{n=8}}& \multicolumn{3}{c}{\textbf{n=10}}\\
        & \textbf{Mistral} & \textbf{Llama3} & \textbf{Qwen} & \textbf{Mistral} & \textbf{Llama3} & \textbf{Qwen2.5} & \textbf{Mistral} & \textbf{Llama3} & \textbf{Qwen2.5} & \textbf{Mistral} & \textbf{Llama3} & \textbf{Qwen2.5}& \textbf{Mistral} & \textbf{Llama3} & \textbf{Qwen2.5}\\
        \midrule
        
        Zero fewshot & 39.13 &  38.31 &  39.73 &  45.28 &  40.76 &  43.52 &  44.53 &  40.55 &  46.49 &  44.01 &  41.11 &  46.53 &  41.50 &  39.15 &  47.24 \\
        1-Fewshot  & 40.43 & 37.54 & 37.94 & 48.28 & 37.35 & 44.69 & 46.11 & 38.27 & 45.07 & 44.85 & 40.40 & 45.46 & 42.39 & 39.94 & 45.05  \\
        2-Fewshot  & 41.56 & 42.73 & 38.98 & 49.29 & 37.77 & 43.16 & 46.69 & 36.83 & 46.02 & 45.35 & 40.54 & 45.03 & 43.53 & 39.35 & 44.86 \\
        3-Fewshot  & 40.70 & 39.41 & 40.68 & 48.60 & 36.90 & 44.27 & 47.43 & 36.94 & 45.59 & 44.83 & 39.97 & 44.89 & 43.08 & 38.87 & 44.77 \\
        4-Fewshot  & 41.58 & 40.96 & 40.99 & 48.77 & 37.41 & 43.86 & 47.59 & 36.72 & 46.19 & 45.56 & 40.55 & 44.64 & 43.72 & 40.06 & 43.37 \\
        5-Fewshot  & 41.17 & 41.72 & 40.36 & 48.50 & 39.39 & 43.65 & 47.70 & 39.27 & 45.81 & 44.94 & 41.27 & 44.55 & 42.31 & 41.07 & 44.21 \\
        \bottomrule
        \end{tabular}%
        }
        \caption{F1 performance scores of different LLMs (\texttt{Mistral-7B}, \texttt{Llama3-8B}, and \texttt{Qwen2.5-7B}) across various retrieval configurations using \textbf{CRAFT}. Results are shown for increasing numbers of retrieved tables ($n = 1, 3, 5, 8, 10 $) and different few-shot prompting strategies. Each row represents a different few-shot variant, with \texttt{zero\_fewshot} indicating zero-shot performance.}
        \label{fig:f1_extended}
    \end{table*}

        \begin{table*}[htbp]
        \centering
    \setlength{\aboverulesep}{0pt}
    \setlength{\belowrulesep}{0pt}
    \setlength{\tabcolsep}{1.5pt}
        \renewcommand{\arraystretch}{1.2}
        \resizebox{\textwidth}{!}{%
        \begin{tabular}{l|ccc|ccc|ccc|ccc|ccc}
        \toprule
        \textbf{Retriever} & \multicolumn{3}{c|}{\textbf{n=1}} & \multicolumn{3}{c|}{\textbf{n=3}} & \multicolumn{3}{c}{\textbf{n=5}} & \multicolumn{3}{c}{\textbf{n=8}}& \multicolumn{3}{c}{\textbf{n=10}}\\
        \cmidrule(lr){2-4}\cmidrule(lr){5-7}\cmidrule(lr){8-10}\cmidrule(lr){11-13}\cmidrule(lr){14-16}
        & \textbf{Mistral} & \textbf{Llama3} & \textbf{Qwen2.5} & \textbf{Mistral} & \textbf{Llama3} & \textbf{Qwen} & \textbf{Mistral} & \textbf{Llama3} & \textbf{Qwen} & \textbf{Mistral} & \textbf{Llama3} & \textbf{Qwen}& \textbf{Mistral} & \textbf{Llama3} & \textbf{Qwen}\\
        \midrule
        \multicolumn{16}{c}{\textbf{Small Models}}\\
        \midrule
        \textbf{Gold} & 43.60 & 41.10 & 52.87 & - & - & - & - & - & - & - & - & - & - & - & - \\
        \midrule
        \textbf{Stage 2(MT)} & 38.83 & 35.58 & 46.49 & 41.71 & 25.78 & 48.37 & 40.54 & 29.11 & 48.97 & 38.09 & 30.38 & 47.76 & 36.96 & 29.53 & 46.67 \\
        \textbf{Stage 2(FT)} & 37.72 & 34.62 & 46.69 & 39.46 & 25.73 & 47.31 & 38.40 & 29.09 & 49.06 & 36.20 & 29.53 & 47.32 & 35.12 & 28.35 & 46.02 \\
        \textbf{Stage 3 Gemini(MT)} & 39.38 & 36.60 & 49.38 & 41.02 & 26.99 & 50.15 & 39.14 & 29.81 & 48.88 & 36.16 & 31.59 & 46.77 & 35.44 & 30.75 & 47.93 \\
        \textbf{Stage 3 Gemini(FT)} & 38.28 & 36.08 & 48.54 & 39.38 & 26.29 & 49.91 & 38.10 & 29.24 & 49.16 & 34.95 & 30.00 & 47.16 & 34.32 & 30.03 & 46.93 \\
        \midrule
        \multicolumn{16}{c}{\textbf{Large Models}} \\
        \midrule
        \textbf{Gold} & 61.38 & 59.30 & 65.31 & - & - & - & - & - & - & - & - & - & - & - & - \\
        \midrule
        \textbf{Stage 2(MT)} & 54.47 & 52.06 & 56.36 & 56.90 & 50.52 & 61.59 & 57.18 & 50.44 & 62.22 & 54.84 & 51.43 & 62.03 & 53.11 & 51.45 & 61.34 \\
        \textbf{Stage 2(FT)} & 54.80 & 51.50 & 56.79 & 57.52 & 50.52 & 60.74 & 58.12 & 50.94 & 62.59 & 55.66 & 51.55 & 61.45 & 53.25 & 51.00 & 61.69 \\
        \textbf{Stage 3 Gemini(MT)} & 55.43 & 53.24 & 59.24 & 59.41 & 51.95 & 63.55 & 56.77 & 50.99 & 63.34 & 54.12 & 51.42 & 63.16 & 52.86 & 50.61 & 62.80 \\
        \textbf{Stage 3 Gemini(FT)} & 55.06 & 53.28 & 58.35 & 59.21 & 51.17 & 62.95 & 57.27 & 51.13 & 62.17 & 54.65 & 51.08 & 61.92 & 53.72 & 50.69 & 60.06 \\
        \bottomrule
        \end{tabular}%
        }
        \caption{F1 performance scores of different LLMs (Top(Small Models): \textit{Mistral-7B-Instruct}, \textit{Llama3-8B-Instruct}, and \textit{Qwen2.5-7B-Instruct}, and  Bottom(Large Models): \textit{Mistral-Small-Instruct}, \textit{Llama-3.3-70B-Instruct}, and \textit{Qwen2.5-72B-Instruct}) across various retrieval configurations using CRAFT for OTT-QA dataset. Results are shown for increasing numbers of retrieved tables ($n = 1, 3, 5, 8, 10 $) and different table retrievals. Each row represents a different table retrieved using different table embeddings, with \textbf{Gold} indicating end-to-end score with the gold table and passage.}
        \label{tab:f1_ottqa}
    \end{table*}

    \begin{figure*}[t]
    \centering
              \includegraphics[width=\linewidth]{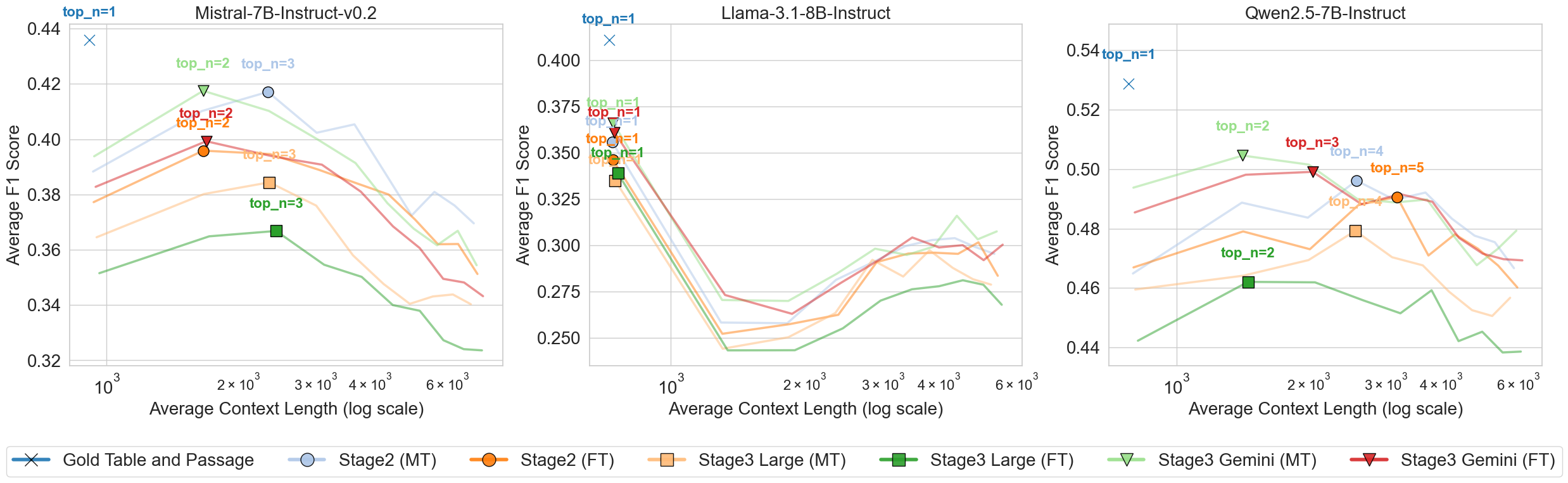}
              \caption{\small End-to-end performance comparison across context lengths with OTTQA dataset. The figure illustrates the relationship between average context length and F1 score for three models(small size models), evaluated over 10 different $ n$-retrieved tables with the OTT-QA datasets.\vspace{-\baselineskip}}
              \label{fig:context_f1_score_ottqa_small}
    \end{figure*} 

    \begin{figure*}[t]
    \centering
              \includegraphics[width=\linewidth]{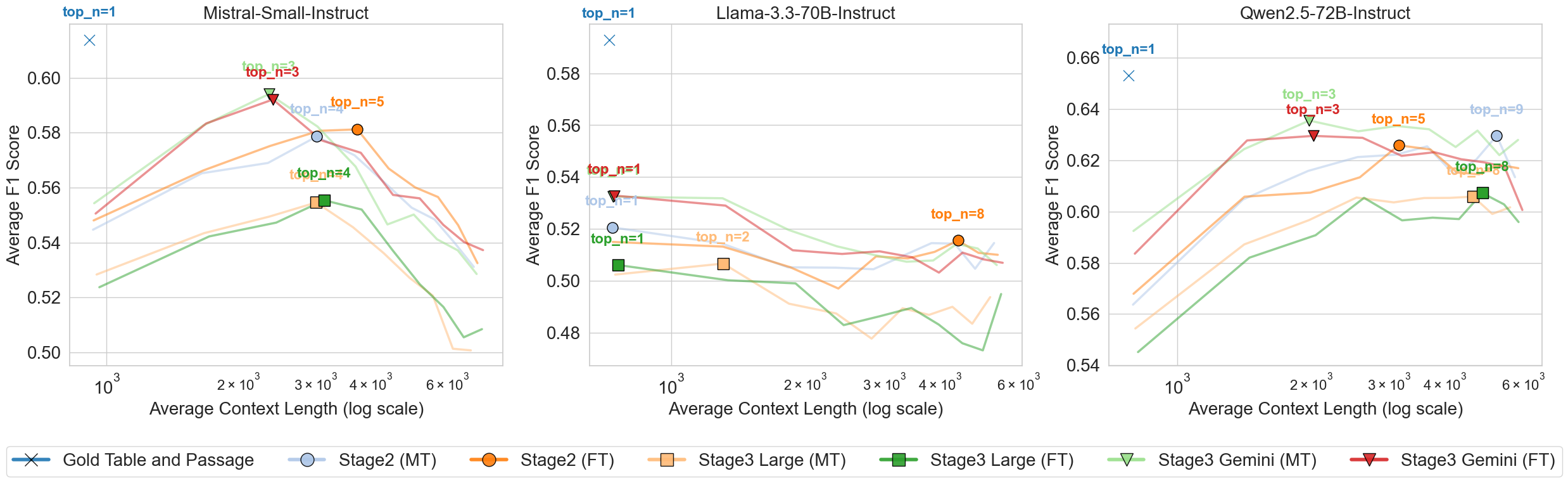}
              \caption{\small End-to-end performance comparison across context lengths with OTT-QA dataset. The figure illustrates the relationship between average context length and F1 score for three models(large size models), evaluated over 10 different $ n$-retrieved tables with the OTT-QA datasets.\vspace{-\baselineskip}}
              \label{fig:context_f1_score_ottqa_large}
    \end{figure*} 

   \paragraph{End-to-End OTT-QA Performance}\label{appdx:end2end-ottqa}
Table~\ref{tab:f1_ottqa} presents the end-to-end F1 performance of different instruction-tuned LLMs on the OTT-QA dataset using our multi-stage retrieval framework. It compares multiple retrieval strategies across varying depths, highlighting how retrieval design impacts downstream QA accuracy.

We consider the following retrieval settings:
\begin{itemize}
\item \textbf{Gold}: Uses the gold table (oracle evaluation setup) paired with gold passages to isolate the effect of table retrieval and provide an upper-bound reference.
\item \textbf{FT (Full Table)}: Uses the top-$n$ retrieved table IDs (from Stage 2 or Stage 3) in their full form as context for answering the question.
\item \textbf{MT (Mini Table)}: Uses compact representations of the retrieved tables (based on table IDs), retaining only top(k=5) rows to reduce noise and improve reasoning efficiency.
\end{itemize}

Since OTT-QA is a complex hybrid multi-hop benchmark that requires reasoning across both tables and passages, we employ an oracle evaluation setup that pairs the retrieved tables with the corresponding gold passages.

Across both small and large models, highlighted by Figure~\ref{fig:context_f1_score_ottqa_small} and Figure~\ref{fig:context_f1_score_ottqa_large} respectively, performance trends reveal a clear trade-off between recall and precision. For smaller $n$, the retrieved tables may not include the gold table, leading to suboptimal results. Conversely, higher $n$ increases the likelihood of retrieving the relevant table but also introduces redundant or noisy tables that dilute reasoning accuracy. Among all settings, Stage 3 (MT) consistently achieves strong performance, particularly for Qwen2.5-7B-Instruct, which reaches its best score at n = 3–5, closely matching the gold-table upper bound. This demonstrates the benefit of the Mini-Table (MT) strategy, which compactly represents relevant content while minimizing contextual noise. For large-scale models such as Qwen2.5-72B-Instruct, Stage 3 Gemini (MT) yields the most balanced results, highlighting how our approach generalizes effectively across model sizes by achieving a ``sweet spot" between retrieval breadth and contextual precision.
    \section{Error Analysis}\label{appdx:error_analysis}
    Our error analysis reveals several consistent patterns in CRAFT’s behavior under imperfect retrieval conditions. We find that retrieved tables often exhibit high topical similarity to the query, yet do not necessarily contain the specific evidence required for multi-hop reasoning, indicating that surface-level relevance is an unreliable proxy for answerability. Moreover, the rank of the gold table does not directly correlate with end-to-end QA performance, therefore even lower-ranked tables can support correct predictions when paired with effective mini-table representations that isolate relevant evidence. We also observe that standard retrieval metrics such as R@1 and R@5 tend to underestimate overall system capability, as downstream reasoning components can compensate for imperfect retrieval by extracting useful signals from non-top-ranked candidates. Finally, using full tables as input introduces substantial noise, frequently leading to missed answers or spurious predictions, whereas mini-table representations reduce irrelevant context and enable more accurate reasoning. Taken together, these findings suggest that improving QA performance requires joint optimization of retrieval and reasoning, with intermediate representations such as mini-tables playing a central role.

    \paragraph{Case Study I: Full Table Fails Despite Correct Retrieval.}
    Table~\ref{tab:case1} presents an example where the gold table (\textit{1997\_AFL\_draft\_1}) is ranked third among the top-5 retrieved tables, yet the full-table approach fails to extract the correct answer, responding with \textit{Pass} (i.e., unable to find relevant information).
    In contrast, the mini-table design successfully identifies \textit{Jamie Elliott} as the answer.
    This illustrates how CRAFT can exhibit low Recall@1 while still producing correct final answers through focused context construction.

    \begin{table*}[!t]
    \centering
    \small
    \setlength{\tabcolsep}{2pt}
    \begin{minipage}[t]{0.48\textwidth}
    \centering
    \begin{tabular}{>{\bfseries}p{3.0cm} p{4cm}}
    \toprule
    Attribute & Value \\
    \midrule
    QID
      & \textit{e4197b9b9bd82bdd} \\[2pt]
    Question
      & Who was the St.\ Kilda pick 11 \\
      & in the 1997 AFL draft? \\[2pt]
    Gold Answer
      & Jamie Elliott \\[2pt]
    Gold Table
      & \textbf{\textit{1997\_AFL\_draft\_1}} \\[2pt]
    Gold Table Rank
      & 3 \\
    \midrule
    \multicolumn{2}{l}{\textbf{Top 5 Retrieved TableIDs}} \\[2pt]
    1 & \textit{1998\_AFL\_Draft\_3} \\[2pt]
    2 & \textit{List\_of\_St\_Kilda\_Football\_Club\_} \\
      & \textit{records\_and\_statistics\_8} \\[2pt]
    \textbf{3} & \textbf{\textit{1997\_AFL\_draft\_1}~\checkmark } \\[2pt]
    4 & \textit{1998\_AFL\_Draft\_2} \\[2pt]
    5 & \textit{List\_of\_St\_Kilda\_Football\_Club\_} \\
      & \textit{records\_and\_statistics\_9} \\
    \midrule
    \multicolumn{2}{l}{\textbf{LLM QA Step}} \\[2pt]
    Full Table & \textit{Pass} (no answer found)~$\times$ \\[2pt]
    Mini-Table & \textbf{Jamie Elliott~\checkmark} \\
    \bottomrule
    \end{tabular}
    \caption{Case Study I: The gold table is ranked third. The full-table approach fails to extract the correct answer, while the mini-table design succeeds by isolating the relevant row.}
    \label{tab:case1}
    \end{minipage}%
    \hfill
    \begin{minipage}[t]{0.48\textwidth}
    \centering
    \begin{tabular}{>{\bfseries}p{3.0cm} p{4cm}}
    \toprule
    Attribute & Value \\
    \midrule
    QID
      & \textit{bde8169851a494cf} \\[2pt]
    Question
      & What album did Elisa's most \\
      & popular original song come out on? \\[2pt]
    \textbf{Gold Answer}
      & \textbf{Asile's World} \\[2pt]
    \textbf{Gold Table}
      & \textit{List\_of\_songs\_recorded\_by\_Elisa\_0} \\[2pt]
    Gold Table Rank
      & 2 \\
    \midrule
    \multicolumn{2}{l}{\textbf{Top-5 Retrieved Table Ids}} \\[2pt]
    1 & \textit{List\_of\_songs\_recorded\_by\_Elisa\_1} \\[2pt]
    \textbf{2} & \textbf{\textit{List\_of\_songs\_recorded\_by\_Elisa\_0} ~\checkmark} \\[2pt]
    3 & \textit{Elisa\_(Italian\_singer)\_1} \\[2pt]
    4 & \textit{Elisa\_(Italian\_singer)\_0} \\[2pt]
    5 & \textit{Raquel\_Olmedo\_1} \\
    \midrule
    \multicolumn{2}{l}{\textit{LLM QA Step}} \\[2pt]
    Full Table & Lotus (2003)~$\times$ \\[2pt]
    Mini-Table & \textbf{Asile's World (2000)}~\checkmark \\
    \bottomrule
    \end{tabular}
    \caption{Case Study II: Even with the gold table ranked second, the full-table approach hallucinates an incorrect answer, while the mini-table approach correctly isolates the relevant information.}
    \label{tab:case2}
    \end{minipage}
\end{table*}
    
    \paragraph{Case Study II: Full Table Hallucination with Closely-ranked Retrieval.}
    Table~\ref{tab:case2} presents a more nuanced failure mode.
    The gold table (\textit{List\_of\_songs\_recorded\_by\_Elisa\_0}) is ranked second, which is a relatively strong retrieval result, yet answer quality diverges sharply between the two approaches.
    The full-table approach returns \textit{Lotus (2003)}, a plausible but incorrect answer drawn from a different row in the retrieved context.
    The mini-table approach correctly returns \textit{Asile's World (2000)} by isolating the relevant row corresponding to Elisa's most popular original song.

    Taken together, these case studies highlight that the benefit of mini-tables is not limited to weak retrieval scenarios.
    Even when relevant tables are retrieved at high ranks, full-table context can introduce sufficient noise through topically similar but irrelevant rows to cause model failure or hallucination.
    Mini-table construction acts as a targeted denoising step, enabling the downstream QA model to focus on the precise evidence needed for multi-hop inference.
    
\end{document}